\RequirePackage[hyphens]{url}
\RequirePackage[RGB]{xcolor}

\documentclass{bmvc2k}

\usepackage{subcaption}
\makeatletter
\let\bmv@makecaption=\@makecaption
\def\@makecaption#1#2{\small{\textcolor{bmv@captioncolor}{\bmv@makecaption{#1}{#2}}}}
\makeatother

\usepackage{xspace}
\usepackage{epsfig}
\usepackage{graphicx}
\usepackage{amsmath, bm}
\usepackage{amssymb}
\usepackage{algorithm}
\usepackage{algorithmic}
\usepackage{mathtools}
\usepackage{array}
\usepackage{multirow}
\usepackage[inline]{enumitem}
\usepackage{multirow}
\usepackage{tabularx}
\usepackage{color}
\usepackage[utf8]{inputenc} 
\usepackage[T1]{fontenc}    
\usepackage{booktabs}       
\usepackage{amsfonts}       
\usepackage{nicefrac}       
\usepackage{microtype}      

\usepackage{tikz}
\usepackage{flushend}

\definecolor{skyscraper}{RGB}{140,140,140}
\definecolor{sky}{RGB}{6,230,230}
\definecolor{building}{RGB}{180,120,120}
\definecolor{tree}{RGB}{4,200,3}
\definecolor{wall}{RGB}{120,120,120}
\definecolor{grass}{RGB}{4,250,7}
\definecolor{sidewalk}{RGB}{235,255,7}
\definecolor{car}{RGB}{0,102,200}
\definecolor{plant}{RGB}{204,255,4}
\definecolor{person}{RGB}{150,5,61}
\definecolor{earth}{RGB}{120,120,70}
\definecolor{fence}{RGB}{255,184,6}
\definecolor{signboard}{RGB}{255,5,153}
\definecolor{pole}{RGB}{51,0,255}
\definecolor{streetlight}{RGB}{0,71,255}
\definecolor{truck}{RGB}{255,0,20}
\definecolor{conveyer}{RGB}{133,0,255}
\definecolor{mountain}{RGB}{143,255,140}
\definecolor{floor}{RGB}{80,50,50}
\definecolor{van}{RGB}{163,255,0}

\DeclareRobustCommand{\tikzcolorbox}[1]{\tikz[overlay]\node[fill=#1,text depth=0pt,text height=6pt,inner sep=2pt, anchor=text,rectangle] {#1};\phantom{#1}}

\makeatletter
\renewcommand{\subsectionautorefname}{\S\@gobble}
\renewcommand{\sectionautorefname}{\S\@gobble}
\makeatother

\makeatletter
\newcommand\Autoref[1]{\@first@ref#1,@}
\def\@throw@dot#1.#2@{#1}
\def\@set@refname#1{
    \edef\@tmp{\getrefbykeydefault{#1}{anchor}{}}%
    \xdef\@tmp{\expandafter\@throw@dot\@tmp.@}%
    \ltx@IfUndefined{\@tmp autorefnameplural}%
         {\def\@refname{\@nameuse{\@tmp autorefname}s}}%
         {\def\@refname{\@nameuse{\@tmp autorefnameplural}}}%
}
\def\@first@ref#1,#2{%
  \ifx#2@\autoref{#1}\let\@nextref\@gobble
  \else%
    \@set@refname{#1}
    \@refname~\ref{#1}
    \let\@nextref\@next@ref
  \fi%
  \@nextref#2%
}
\def\@next@ref#1,#2{%
   \ifx#2@ and~\ref{#1}\let\@nextref\@gobble
   \else, \ref{#1}
   \fi%
   \@nextref#2%
}
\makeatother

\title{Semantically-Aware Attentive Neural Embeddings for Long-Term 2D Visual Localization}

\makeatletter
\DeclareRobustCommand\onedot{\futurelet\@let@token\@onedot}
\def\@onedot{\ifx\@let@token.\else.\null\fi\xspace}
\makeatother
\def\etal{\emph{et al}\onedot}

\def\ie{\emph{i.e}\onedot}
\def\eg{\emph{e.g}\onedot}

\def\R{\mathbb{R}}

\newcommand{\mat}[1]{\bm{#1}}

\def\mF{\mat{F}}
\def\mM{\mat{M}}
\def\mW{\mat{W}}

\def\conv{\circledast}

\definecolor{redcol}{rgb}{1, 0, 0}
\definecolor{bluecol}{rgb}{0, 0, 1}

\renewcommand{\paragraph}[1]{\smallskip\noindent{\bf{#1}}}

\addauthor{Zachary Seymour}{}{1}
\addauthor{Karan Sikka}{}{1}
\addauthor{Han-Pang Chiu}{}{1}
\addauthor{Supun Samarasekera}{}{1}
\addauthor{Rakesh Kumar}{}{1}

\addinstitution{
 SRI International\\
 201 Washington Road\\
 Princeton, NJ
}

\runninghead{Seymour, \etal}{SAANE for 2D Visual Localization}

\begin{document}

\maketitle

\begin{abstract}

We present an approach that combines appearance and semantic information for 2D image-based localization (2D-VL) across
large perceptual changes and time lags.  Compared to appearance features, the semantic layout of a scene is generally
more invariant to appearance variations.  We use this intuition and propose a novel end-to-end deep attention-based
framework that utilizes multimodal cues to generate robust embeddings for 2D-VL. The proposed attention module predicts
a shared channel attention and modality-specific spatial attentions to guide the embeddings to focus on more reliable
image regions.
We evaluate our model against state-of-the-art (SOTA) methods on three challenging localization datasets. We report an
average (absolute) improvement of $19\%$ over current SOTA for 2D-VL.  Furthermore, we present an extensive study
demonstrating the contribution of each component of our model, showing $8$--$15\%$ and $4\%$ improvement from adding
semantic information and our proposed attention module. We finally show the predicted attention maps to offer useful insights into our model.

\end{abstract}

\section{Introduction}

Visual localization (VL) is the problem of estimating the precise location of a captured image and is
crucial to applications in autonomous navigation~\cite{se2002mobile, kendall2015posenet, lim2012real,
	sattler17:are-3d-models-necessary, naseer17:semantics-aware-visual-localization, maddern20171,
	chowdhary2013gps}. Here, we target the problem of \emph{long-term
	VL}~\cite{stenbord18:localization-semantically-segmented, toft2017long,
naseer17:semantics-aware-visual-localization, maddern20171}  in real-world environments, which is
required to operate under (1) extreme perceptual changes such as weather and illumination, and (2) dynamic scene changes and
occlusions such as moving vehicles. We also evaluate this task to focus on a key requirement of long-term VL, which is
to localize within a narrow radius of ${<}10$m instead of $25$m-- the localization radius for evaluating the
related task of \emph{place recognition}~\cite{arandjelovic18:netvlad, gomez-ojeda15:appearance-invariant}.

Prior works on VL generally utilize two broad classes of methods: 3D structure-based localization
(3D-VL) and 2D image-based localization (2D-VL)~\cite{sattler17:are-3d-models-necessary,sattler2018benchmarking}, both
of which utilize
local or global low-level appearance descriptors.
3D-VL methods typically associate a local descriptor with each point in a 3D model of the scene~\cite{irschara2009structure,li2012worldwide,sattler2015hyperpoints}, while 2D-VL methods extract a holistic descriptor or multiple local descriptors for query-based matching~\cite{arandjelovic14:disloc,torii15:repetitive-structures,arandjelovic18:netvlad,torii18:dense-vlad}.
A primary drawback with methods using low-level features is that they are not robust to changes in viewing conditions~\cite{schonberger2018semantic,naseer17:semantics-aware-visual-localization}; while 3D models are less scalable and present greater computational complexity.
To this end, we focus on improving 2D-VL methods specifically for operating under severe changes in viewing conditions in large-scale urban environments and over large time lags~\cite{maddern20171}.   

\begin{figure}[t]
	\centering
	\includegraphics[width=0.7\linewidth]{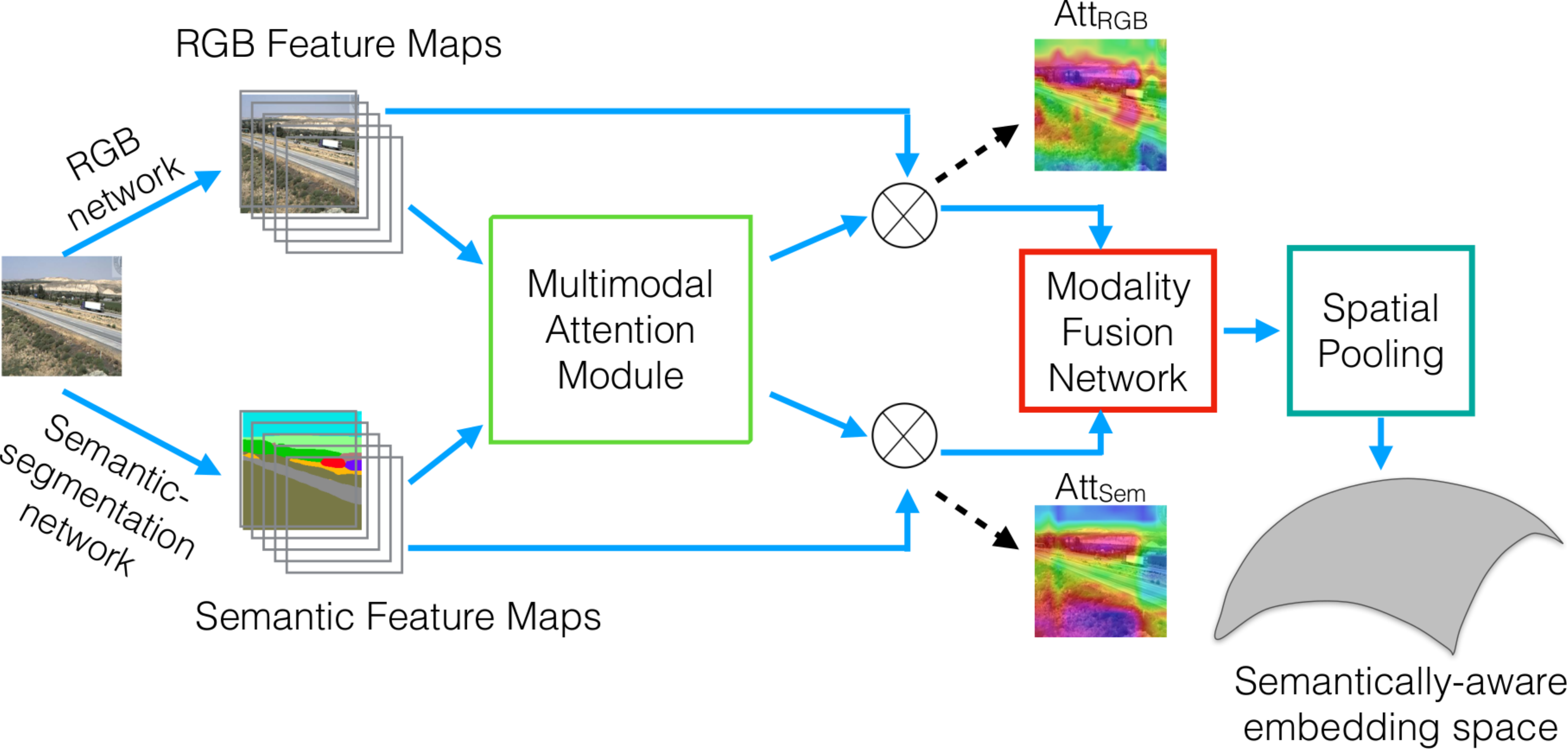}
	\caption{\small{We show a high-level overview (top) of our approach (SAANE) that learns to effectively combine
		appearance and semantic information with modality-specific spatial attention to compute
semantically-aware feature embeddings in the presence of severe changes in viewing conditions.}}
	\label{fig:intro}
\end{figure}

Recently 2D-VL methods have used deep convolutional neural networks (DCNN) based
frameworks that learn to generate view-invariant image representations (or embeddings)
~\cite{gomez-ojeda15:appearance-invariant, arandjelovic18:netvlad, zhu18:apa}. Despite advancing the
state-of-the-art, DCNN still suffer from two key issues in long-term VL.  First, due to the reliance on low or
mid-level appearance features, DCNN methods suffer a loss in accuracy for large changes in viewing conditions~\cite{naseer17:semantics-aware-visual-localization,
schonberger2018semantic}.  Second, since DCNN methods extract holistic representations from an entire image without any
explicit understanding of scene composition, the resulting image representation is degraded by non-discriminative
visual elements (vehicles)~\cite{naseer17:semantics-aware-visual-localization, zhu18:apa, piasco2018survey}.
To alleviate these issues, we propose a novel DCNN framework to improve the robustness of learned visual embeddings by incorporating
\begin{enumerate*}[label=(\roman*)]
\item high-level semantic information inside the neural network, and
\item an attention-based framework that
uses both appearance and semantic information to guide the model to focus on informative and stable
image regions.
\end{enumerate*}

Our method, referred to as Semantically-Aware Attentive Neural Embeddings (SAANE) (see \Autoref{fig:intro}), is an
end-to-end trainable deep neural network, that effectively combines semantic\footnote{By \emph{semantic} we refer to
representations that provide high-level information about the input; \eg, per-pixel depth or semantic segmentation
maps.} and mid-level appearance features with spatial attention to learn semantically-aware embeddings.  The
design of SAANE is inspired by recent advances in deep multimodal and attention guided learning
\cite{ahuja2018understanding, woo2018cbam, xu2016ask, yang2016stacked, afouras2018deep} and is composed of three key
modules:
\begin{enumerate*}[label=(\roman*)]
\item the \textit{modality fusion module:} to fuse information from the two modalities,
\item the \textit{multimodal attention module:} to use the fused representation to predict a shared channel attention
	and separate modality-specific spatial attentions, and
\item the \textit{spatial pooling module:} to pool the attended multimodal features to generate embeddings,
	which are then used to retrieve matches for a query image.
\end{enumerate*}
We use convolutional feature maps from models pre-trained for image classification and semantic segmentation as
appearance and semantic representations respectively.
Our model is trained end-to-end by using a ranking based objective.

The key motivation behind our approach is that compared to low/mid-level appearance descriptors, the spatial layout
of semantic classes in the image yields scene descriptions that have a higher invariance to large changes in viewing
conditions for long-term VL. The semantic understanding of the image content along with spatial attention helps
to determine which regions of the scene may be unreliable for localization across large time scales.
A key advantage of SAANE over prior works, that have attempted to improve the robusness of low-level features,
is that it offers a principled end-to-end module to learn semantically-aware representations with
spatial attention- guided by multimodal features.

The key contributions of this paper are as follows:
\begin{enumerate}
\item We incorporate higher-level semantic information along with commonly-used mid-level appearance features to enhance
	the quality of the \emph{learned} image embeddings for visual localization under large appearance changes.

\item We propose an attention-based neural module to allow the model to focus on stable and discriminative regions. To the best	of our	knowledge, ours is the first work to propose a DCNN-based pipeline that combines multimodal representations (appearance and semantic), with spatial attention for visual localization.

\item We perform extensive ablation studies across three datasets to show the
	contributions of each part of our model. We particularly investigate and establish the necessity of the proposed separable
	spatial with shared channel attention for current problem.

\end{enumerate}

\section{Related Work}

Methods for image-based visual localization generally fall into two classes: 3D structure-based localization
(3D-VL)~\cite{sattler2015hyperpoints,zeisl2015camera,svarm2017city,sattler17:are-3d-models-necessary,schonberger2018semantic}
and 2D image-based localization (2D-VL)~\cite{cummins2008fab,
cummins11:fab-map,torii18:dense-vlad,arandjelovic18:netvlad,zhu18:apa}. 3D-VL methods create a 3D model of the scene by
either using Structure-From-Motion (SfM) or associating local patches to 3D point clouds.  On the other hand, 2D-VL
methods formulate visual localization as an image retrieval problem by matching a query image with geo-tagged database
images for approximate localization.  
In this paper we focus on improving 2D-VL methods in large-scale scenes, which can then be used in combination with 3D
methods for more precise localization as shown in~\cite{sattler17:are-3d-models-necessary}.

Most initial works for visual localization relied on matching Bag of Visual Word (BoVW)-type
features~\cite{glover10:fabmapratslam, arandjelovic14:disloc} or using global image descriptors in addition to
sequential search~\cite{milford12:seqslam}.  Building on the success of deep convolutional neural networks (DCNN) in
other areas~\cite{girshick2015fast, krizhevsky2012imagenet}, recent works have extensively demonstrated the
effectiveness of off-the-shelf DCNN features for visual place recognition and VL
~\cite{sunderhauf15:cnn-placerecog,chen14:cnn-placerecog,garg18:lost}.  Several works have also focused on
improving feature pooling methods for off-the-shelf features~\cite{babenko2015aggregating,
yue2015exploiting,torii18:dense-vlad} or those learned in end-to-end pipelines~\cite{arandjelovic18:netvlad,
zhu18:apa}.

In general, (deep) learned global descriptors are more robust than hand-crafted features but are still susceptible to perceptual aliasing from repeated patterns such as road markings and buildings that introduce indistinguishable global matches; \ie, only specific, meaningful regions are useful for localization~\cite{chen17:only-look-once,tolias2015particular,finn2016deep}. 
In prior works, the contribution of combining mid-level image features with higher-level semantic information has been entangled with the addition of 3D information~\cite{schonberger2018semantic,radwan2018vlocnet++}.
Semantics were also used to focus on manually-selected regions~\cite{naseer17:semantics-aware-visual-localization} or to enhance off-the-shelf appearance features~\cite{arandjelovic2014visual,garg18:lost}.
Our method's novel multimodal attention module learns not only to combine features across modalities but also to focus on the most stable regions by discovering region types in a data-driven fashion. 

We also situate our work within the space of making image features semantically-grounded and more robust to appearance changes. 
A typical approach to matching image features across large perceptual changes is to learn a transformation from one type of appearance to another---such as across seasons~\cite{naseer14:robust-visual-localization} or across different times of day~\cite{lowry14:morning-to-afternoon,anoosheh2018night}---or to train a DCNN for appearance invariance using paired images~\cite{gomez-ojeda15:appearance-invariant,chen17:deep-learning-place-recog,chen2018learning}.
Our method seeks to integrate this innovation with stable semantic scene features to further improve the invariance of these features.

Our work is also related to recent works on combining vision and language for the Visual Question Answering (VQA) task~\cite{antol2015vqa}. 
The underlying fusion networks vary from pooling methods such as sum/bilinear pooling~\cite{fukui2016multimodal} to more complex attention based methods~\cite{teney2018tips, anderson2018bottom, yang2016stacked, lu2016hierarchical}. 
Although the proposed multimodal attenion module is motivated by these works, it is both different and better suited to
the requirements of the long-term VL task, which is fundamentally different from the VQA task. For example, the
appearance and semantic modalities in VL are more
closely-aligned as compared to vision and language in VQA, due to which additional blocks are required for 
aligning the modalities prior to fusion.

\def\fmid{$App_\text{mid}$}
\def\fhigh{$Sem_\text{high}$}
\section{Approach: SAANE}

We now describe our approach, refered to as \textbf{SAANE} (Semantically-Aware Attentive Neural Embeddings), in detail.
Our model passes an input RGB through two DCNNs pre-trained for image classification and semantic segmentation to
obtain mid-level appearance features (denoted as \fmid) and high-level semantic features (denoted as \fhigh)
respectively.  We denote feature maps from layer $l$ of the image-classification and semantic-segmentation CNNs as
$\mF^{A}_{l} \in \R^{C_A \times H \times W}$ and $\mF^{S}_{l} \in \R^{C_S \times H \times W}$ respectively, where $C_A$
and $C_S$ are the number of channels. We drop the notation $l$ in the rest of the paper for brevity. We work in a
supervised setting with a database of geo-tagged images captured under different viewing conditions.

SAANE is an end-to-end trainable DCNN that learns to generate robust image embeddings that, in addition to being
invariant to changes in viewing conditions, are aware of semantic composition of the scene and focus explicitly on
informative visual elements due to the use of spatial attention.  Our model (\autoref{fig:intro}) consists of three
NN-based modules: \begin{enumerate*}[label=(\roman*)] \item  the modality fusion module, \item the multimodal attention
(MM-Att) module, and \item  the spatial pooling module.  \end{enumerate*} SAANE operates by first transforming and
fusing features from the appearance and the semantic input streams by using the modality fusion module. The output is a
semantically-informed multimodal representation of input, which is then used to estimate per-modality spatial attentions
using the proposed MM-Att module.  We encourage sharing of information between the two modalities by building upon a prior work for unimodal attention~\cite{woo2018cbam} and computing a shared channel attention which is then used to generate separate spatial attentions as described below.
The output from this module is used to refine the feature maps from both modalities, which are then fused together with
another modality fusion module. Finally, we use a spatial pooling to output the embeddings. 
We train our model in an end-to-end fashion to learn each of these modules for visual localization. We now describe these modules along with the loss function and the training procedure.

\paragraph{Modality Fusion Module.} The modality fusion module aligns the feature maps---\fmid{} and
\fhigh{}---by first projecting them in a common space and then adding them together~\cite{yang2016stacked,
ahuja2018understanding}. We use $1 \times 1$ convolutions, denoted by $\mW_{A}^{1} \in \R^{C \times C_A \times 1 \times 1}$
and $\mW_{S}^{1} \in \R^{C \times C_S \times 1 \times 1}$ for the appearance and semantic streams respectively, to project the feature maps in a $C$-dimensional common space.
	\begin{equation}
	\begin{aligned}
		\mF^{M} &= \mW_{A}^{1} \conv \mF^A +\mW_{S}^{1} \conv \mF^S \\
		{} &= \mF^M_A + \mF^M_S,
	\end{aligned}
	\label{eq:initial_projection}
	\end{equation}

	where $\mF^{M}$ is the fused multimodal representation of the image, $F^{M}_{A}$ and $F^{M}_{S}$ are the aligned features maps from \fmid{} and \fhigh{} respectively, and
	$\conv$ is the convolutional operator.  The output is a semantically-informed multimodal representation of the
	input and is used as input to both MM-Att and later to the spatial pooling module. Although recent methods have
	used sophisticated pooling approaches~\cite{fukui2016multimodal}, we opted for
	projected sum pooling, as it uses few trainable parameters and maintains the spatial
	configuration of the feature maps, as required for the attention step.

\paragraph{Multimodal Attention Module (MM-Att).} The multimodal attention module is responsible
for predicting attention at different spatial locations independently for appearance and semantic input streams. The
spatial attention allows our network to selectively focus on discriminative and stable visual elements such as buildings
instead of confusing/dynamic elements such as cars/pedestrians \cite{zhu18:apa}.  This results in
embeddings that are more robust to perceptual changes especially in urban environments. We use the combined multimodal representation, computed by the fusion
module, to sequentially predict a shared channel attention (denoted by $\mM_c \in \R^{C}$) and individual spatial
attentions for the two modalities (denoted by $\mM_{xy}^A \in \R^{H \times W}$ and $\mM_{xy}^S \in \R^{H \times W}$ for
appearance and semantic channels respectively). We believe that a tied channel attention allows sharing of information
between the two modalities leading to a better spatial attention (as also evident later in our results in
\autoref{sec:exp}). The channel attention is computed by summarizing the feature maps
across the spatial dimensions by \textit{average} ($\mF^{M}_\text{avg}$) and \textit{max} ($\mF^{M}_\text{max}$) pooling, and passing them through a multi-layer perceptron (MLP) followed by an addition and a non-linearity:
		\begin{equation}
			\mM_c = \sigma(\phi(\mF^{M}_\text{avg}) + \phi(\mF^{M}_\text{max})),
		\end{equation}
where $\sigma$ denotes the sigmoid function, $\phi$ denotes a two-layer MLP shared across the two pooled inputs.
The refined multimodal representation with attended channels is computed as
$\hat{\mF}^{M} = \mF \odot \mM_c$,
where $\odot$ denotes element-wise multiplication with appropriate broadcasting (copying) of attention values along the spatial dimension.

The refined image representation is then used to predict per modality spatial attentions
		by using two $7 \times 7$ convolutional filters---${\mW}_A^{2}$ and $\mW_S^{2}$---for appearance and semantic
		input streams, respectively. $\hat{\mF}^{M}$ is pooled across the channel
		dimension by both \textit{average} ($\hat{\mF}^{M}_\text{avg}$) and \textit{max}
		($\hat{\mF}^{M}_\text{max}$) pooling and concatenated across the channel dimension and convolved with
		the corresponding filters.
The spatial attention maps are then used with the common channel attention to attend to the transformed maps from \autoref{eq:initial_projection} and generate refined features denoted as $\hat{\mF}^A$ and $\hat{\mF}^S$ for \fmid{} and \fhigh{}, respectively:
\begin{gather}
			M_{xy}^{Z} = \mM_c \odot \sigma(\hat{\mW_Z^{2}} \conv
			([\hat{\mF}^{M}_\text{avg}\,;\hat{\mF}^{M}_\text{max}])) \quad \forall Z \in \{A, S\}
		\end{gather}
The final attended features $\hat{\mF}^{A}$ and $\hat{\mF}^{S}$ from the appearance and semantic input streams	are given by
$\hat{\mF}^A = \mF^M_A \odot  M_{xy}^{A}$ and $\hat{\mF}^S = \mF^M_S \odot  M_{xy}^{S}$ respectively.
 We use another modality fusion module to fuse these refined features and then input them
		to the spatial pooling module.

\paragraph{Spatial Pooling Module.} This module is responsible for pooling the information from the attended and fused
features from the previous modules. In this work we use spatial pyramid pooling
(SPP)~\cite{lazebnik06:spatial-pyramid-matching} since it has been previously shown to be effective, and does not include
any trainable parameters.
Other equally effective alternatives, such as NetVLAD~\cite{arandjelovic18:netvlad}, would work in our framework.
Finally, following the intuition of Ranjan \etal~\cite{ranjan2017l2}, we $L_2$-normalize these embeddings and scale them by a factor of $\alpha = 10$.

	\paragraph{Loss.}
    We use a max-margin-based triplet ranking loss function to learn our model~\cite{schroff2015facenet}. This loss
    optimizes the network such that images from similar locations should be located closer in the embedding space than images from different locations.
    For computational efficiency, we form triplets in an online manner by sampling them from each minibatch~\cite{schroff2015facenet}.

\section{Experimental Results}
\label{sec:exp}

We first describe our train/test  datasets along with the evaluation metric (\autoref{sec:datasets}).
We then compare our model against a strong DCNN baseline while carefully demonstrating the contribution of each component of our model (\autoref{sec:quant}).
Thereafter, we discuss qualitative results to provide important
insights into the proposed attention module (\autoref{sec:qual}).
We finally compare our approach with state-of-the-art (SOTA) 2D-VL methods on the three test datasets (\autoref{sec:sota}).

\begin{table}[t]
 	\centering
\resizebox{0.9\linewidth}{!}{%
 \begin{tabular}{lcccccccc}
        \toprule
		\multirow{2}{*}{Method}	& \multirow{2}{*}{Semantic}  & \multirow{2}{*}{Attention} & \multirow{2}{2cm}{\centering Nordland Sum.$\to$Win.} & \multicolumn{2}{c}{St. Lucia} & \multicolumn{2}{c}{RobotCar} \\
		& &  &  & Average & Worst & AM$\to$PM & Sum.$\to$Win. \\ \midrule
		App	 [Baseline] & & & 58.0 & 71.4 & 55.5 & 36.6 & 79.6 \\ \midrule
		App+Sem    & \checkmark & & 74.6 & 74.9  & 60.6 & 61.4 & 87.9\\
		App-Att	   & & \checkmark & 74.2 & 74.0 & 60.3 & 60.9 & 86.5 \\
		App-Att+Sem-Att & \checkmark & \checkmark & 68.5 & 73.9 & 58.9 & 62.9 & 83.6\\
		SAANE [Proposed]     & \checkmark & \checkmark & \textbf{77.3} & \textbf{78.3} & \textbf{69.1} &	\textbf{67.3} & \textbf{88.5}    \\ \midrule
		DenseVLAD~\cite{torii18:dense-vlad}   &  &  & 19.2 & 64.5 & 35.5 & 22.9 & 78.6  \\
		NetVLAD~\cite{arandjelovic18:netvlad} &   &  & 36.7 & 57.7 & 33.1 & 37.0 & 86.8  \\
		AMOSNet$^*$~\cite{chen17:deep-learning-place-recog} &  &  & 45.6 & 75.3  & 63.7 & 45.3 & 75.1  \\
		APANet$^*$~\cite{zhu18:apa}	& 		& $\dagger$  & 27.0 & 42.4 & 26.1 & 11.2 & 58.2 \\
      APANet-MM$^*$~\cite{zhu18:apa}	& 	\checkmark	&  $\dagger$  & 24.9 & 41.2 & 23.0 & 16.4 & 58.3  \\ \bottomrule
    \end{tabular}}
    \vspace{0.55em}
    \caption{\small{Evaluation results of the different models discussed in \autoref{sec:quant} and \autoref{sec:sota}. We
	report area under the precision-recall curve (AUC) for each method. We show results with prior state-of-the-art
	models below the double line.  $^*$ denotes models that we implemented and trained on SPED dataset that was used
	to train our model. $\dagger$ denotes variants of spatial attention that are distinct from our proposed
method.}}
    \label{tab:saane}
    \end{table}
\subsection{Datasets and Evaluation}
\label{sec:datasets}

\begin{table}
\centering
\resizebox{0.8\linewidth}{!}{%
\begin{tabular}{lccccc}
\toprule
 & \multicolumn{2}{c}{\#Frames} & \multicolumn{3}{c}{Appearance Variation} \\
Dataset & Database & Query & Viewpoint & Illumination & Weather \\ \midrule
Nordland~\cite{neubert2015superpixel} & 1403 & 1403 & None & Yes & Yes \\
St. Lucia~\cite{glover10:fabmapratslam} & 1409 & $1350\times 9^*$ & Slight & Yes & No \\
RobotCar~\cite{maddern20171} AM$\to$PM & 2210 & 2254 & Slight & Yes & Slight \\
RobotCar~\cite{maddern20171} Sum.$\to$Win. & 2065 & 2408 & Slight & Slight & Yes \\ \bottomrule
\end{tabular}}
\caption{A summary of the size of each of our test datasets and the variations they contain. ($^*$ Following standard practice, results on the St. Lucia dataset are averaged over 9 different query sets from different times of day.)}
\label{tab:test_data}
\end{table}

\paragraph{Datasets.} We utilize a version of the Specific Places Dataset (SPED)
~\cite{chen17:deep-learning-place-recog} to train our model.  We randomly sample $\sim 2600$ cameras from the Archive of
Many Outdoor Scenes~\cite{jacobs2007consistent} and download images collected every half hour from Feb-Aug 2014.  We
remove all images where (i) the camera feed was corrupted, obscured, or too dark for visibility, and (ii) capture
location of the camera was not fixed.
The final dataset comprises $1.3$ million images drawn from $2079$ cameras featuring significant scene diversity,
ranging from urban roads to unpopulated landscapes, and appearance changes due to seasonal and day-night cycles.  We
train our model on this dataset and evaluate the performance on 2D-VL on three challenging datasets as shown in
\autoref{tab:test_data}).  Since our focus is long-term VL and not place recognition as mentioned previously, we use the
benchmarks as described in~\cite{chen17:deep-learning-place-recog, sunderhauf15:cnn-placerecog} .  For St. Lucia and
Nordland, we use the procedure in~\cite{sunderhauf15:cnn-placerecog} due to insufficient experimental details
in~\cite{chen17:deep-learning-place-recog}.  Since the remaining datasets used
in~\cite{chen17:deep-learning-place-recog} were either less challenging (Eynsham) or small (Gardens Point), we adapt
RobotCar~\cite{maddern20171} to provide a more challenging test-bed.  We refer readers to \autoref{appsec:ds} for
additional details regarding the datasets.

\paragraph{Evaluation Metric.} We report Area Under the Curve (AUC) by constructing precision-recall
curves using ``the ratio test''~\cite{sunderhauf15:cnn-placerecog, chen17:deep-learning-place-recog}.  In brief, a match between a query image and the
database is considered positive if the ratio of the Euclidean distances of the best match and the second-best match by
nearest neighbors is above some threshold.
A positive match is a true positive if it is within $5$ frames of the ground truth.\footnote{For
Nordland, we use the synchronized frame correspondence. On the other test datasets, $5$ frames covers the ground-truth
frame variance ($0$--$25$ m), a stricter requirement for positive localization than is typically used for place
recognition ($25$--$40$ m). See Appendix A for a detailed discussion.} A precision-recall curve is then constructed by
varying the threshold on the ratio test.
The use of the ratio test enables the measurement of the frame-by-frame matches without relying on the specific pose being made available, which is the case for our test datasets (collected as synchronized traversals of the same route).
More concretely, as the ratio test requires the feature distance between the best and second-best match to be above some
threshold, it actually provides a stricter requirement for correct localization than Recall@1 metric, which is used to evaluate place
recognition.

\paragraph{Prior Methods.} We refer readers to \autoref{appsec:impl} for implementation details regarding our model. For a fair
comparison with prior methods, we use the same backbone DCNN networks for all methods.  We implement
AMOSNet~\cite{chen17:deep-learning-place-recog} by fine-tuning all layers of the DCNN on the SPED dataset.
We also implement a recent SOTA method using both attention and pyramid pooling: Attention-based Pyramid Aggregation
Network (APANet)~\cite{zhu18:apa}. We first implement their proposed cascade pyramid attention block using the
appearance features.  Although their method did not utilize semantic information, we also experiment with a stronger
counterpart that uses two cascade pyramid attention blocks over the multimodal feature maps as used in our work
(plus our multimodal projection as in \autoref{eq:initial_projection}).  We use the learned attention to sum pool the spatial pyramid features across both modalities to achieve the final embedding. 
 We also compare with
implementations of DenseVLAD~\cite{torii18:dense-vlad} and NetVLAD~\cite{arandjelovic18:netvlad}, trained on the
Pitts30k visual place recognition dataset~\cite{torii15:repetitive-structures}, provided by the respective authors.

\subsection{Quantitative Results}
\label{sec:quant}
The performance of the proposed method along with different baselines and prior SOTA approaches is presented
in~\autoref{tab:saane}. We first compare our model with different baselines to highlight the benefits of the proposed
ideas in using
\begin{enumerate*}[label=(\roman*)]
	\item \textit{semantic} information, and
	\item novel \textit{multimodal attention module} to focus on discriminative regions for visual localization.
\end{enumerate*}

We validate the benefits of semantic information by comparing the performance of the baseline model using only
appearance information (App [Baseline]) with that using both appearance and semantic information
(App+Sem). Across our test datasets, there is an average absolute improvement of 
$9\%$, with the largest gain ($25\%$) on the RobotCar AM$\to$PM by using semantic information.  

The remaining variants (App-Att, App-Att+Sem-Att, and SAANE) serve to demonstrate the benefits of the
proposed \emph{attention} module.  Between App [Baseline] and App-Att, we see an average performance hike of $8\%$
(averaged on all datasets) from the attention module that learns to focus on informative visual regions.  The improvements on
Nordland ($16\%$) from attention alone demonstrate the ability of the module to suppresses confusing regions.
These consistent gains demonstrate the benefits of using spatial attention with appearance information.
However, this does not seem to be the case if we naively
compute attention for both modalities with separate attention modules and then combine the resulting features. 
For example, while comparing App-Att+Sem-Att---the model variant that predicts separate attention over each modality---with App+Sem, we observe only minor improvements on the RobotCar AM$\to$PM dataset ($2\%$), 
likely because there is no sharing of information between the two modalities to encourage semantically informed and
more consistent attention maps.

Our model addresses this by using
the fused multimodal image representation to predict spatial attention for each modality by first predicting an intermediate \emph{shared channel attention}.  
SAANE yields the best performance across all variants on each dataset ($12\%$ improvement
over the baseline and $4\%$ over App+Sem and $5\%$ over App-Att+Sem-Att).  Both Nordland ($9\%$) and St.
Lucia ($11\%$ in the worst case) are further refined by sharing channel attention across modalities, while the most
perceptually-challenging test, RobotCar AM$\to$PM, sees a further performance increase of $4\%$ over
App-Att+Sem-Att and of $31\%$ over App [Baseline]. The proposed model is thus able to show consistent improvements over different
baselines across all datasets for the task of visual localization.

\subsection{Qualitative Results}
\label{sec:qual}
\begin{figure}[t]
	\centering
\includegraphics[width=0.8\linewidth]{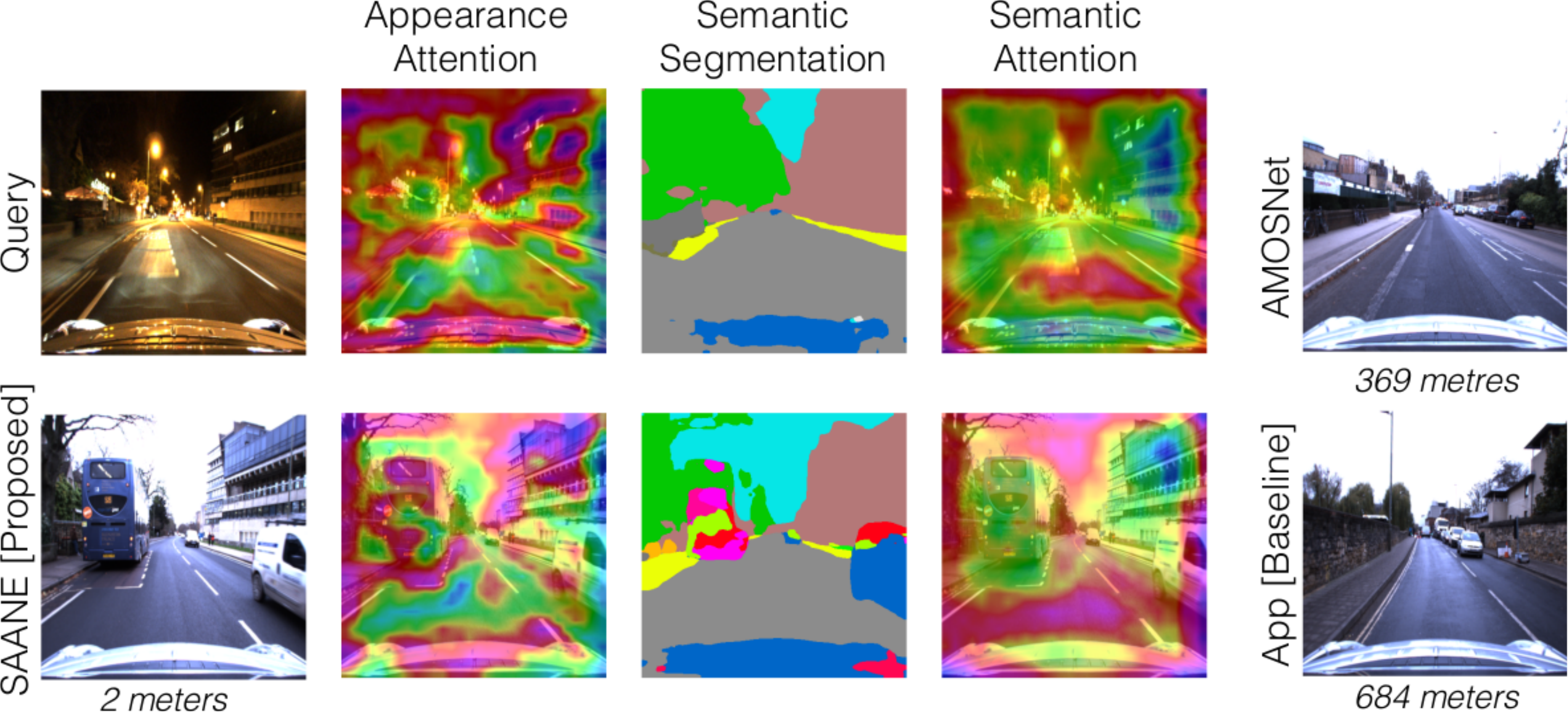}

\caption{\small{We present retrieval results  under challenging appearance variations from RobotCar
	AM$\to$PM to evaluate the quality of the our model and visualize the attention maps. We show
	a query image, database image retrieved using different methods, predicted attention maps, and the pre-computed
	semantic segmentation. The figure highlights that our model is able to retrieve the correct sample with
	and also produces attention maps that are consistent across extreme appearance changes. The palette for semantic
	segmentation is derived from ADE20K, as follows: \tikzcolorbox{skyscraper}, \tikzcolorbox{sky},
	\tikzcolorbox{building}, \tikzcolorbox{tree}, \tikzcolorbox{wall}, \tikzcolorbox{grass},
	\tikzcolorbox{sidewalk}, \tikzcolorbox{car}, \tikzcolorbox{plant}, \tikzcolorbox{person}, \tikzcolorbox{earth},
	\tikzcolorbox{fence}, \tikzcolorbox{signboard}, \tikzcolorbox{pole}, \tikzcolorbox{streetlight},
	\tikzcolorbox{truck}, \tikzcolorbox{conveyer}, \tikzcolorbox{mountain}, \tikzcolorbox{floor},
	\tikzcolorbox{van}. Figure best viewed in color.}} \label{fig:rbc}
\end{figure}

We show top retrievals in \autoref{fig:rbc} for a case with significant variations in viewing conditions.
We visualize the attended regions by showing the spatial attention maps from both modalities along with the semantic segmentation maps.
We show a query image from the Night Autumn matched against retrieved database images from the Overcast Autumn set of RobotCar (\ie, AM$\to$PM in \autoref{tab:saane}).
For methods relying only on appearance information (\ie, AMOSNet and the baseline), the retrieved images in the rightmost column are incorrect but present similar scene layouts, while our model retrieves a match within two meters of the query location.
We see that across time of day, the maps from both attention modalities remain consistent and focus on stable features,
with the appearance attention focusing mostly on fine details such as lane markings or architectural elements and semantic attention giving emphasis to scene layout and skyline shape.
Interestingly, we note the bus present in the matched database image, obscuring a significant part of the scene.
While the appearance modality attends highly to the bus's features, as if it were any other structure, we can see that the semantic attention module has
learned to disregard this region as a dynamic object and gives more emphasis to the remaining scene layout.
These results show that the proposed attention module guides the features to look at consistent regions even across extreme changes in viewing conditions.
We refer readers to \autoref{appsec:cat} for further analysis on the semantic-classes of regions attended across both
modalities. 

\subsection{Comparison with state-of-the-art} \label{sec:sota} \autoref{tab:saane} shows the comparison of our model
with several SOTA techniques.
Our model shows consistent improvements on the test datasets in comparison to DenseVLAD ($28\%$ average absolute) and NetVLAD ($19\%$), both SOTA 2D-VL methods.
We note that NetVLAD performs comparably to our model on RobotCar Sum.$\to$Win. where the appearance changes are relatively minor; however, its performance is much lower on the test datasets with more extreme changes, which is consistent with prior work~\cite{schonberger2018semantic,garg18:lost}.
We also observe an average absolute improvement of $12\%$ (across all datasets) over AMOSNet, which was the previous SOTA method on both the Nordland and St. Lucia datasets.
We note that for datasets which present minor appearance variation (\eg, St. Lucia), nearly the same result is achieved from fine-tuning on SPED (AMOSNet) as from adding additional semantic features (App+Sem).
However, the capacity of our complete (SAANE) attention module to further refine the localization is shown again, with $3\%$ improvement over AMOSNet.
For cases with more severe appearance variation, we see a much larger improvement from training the proposed modules combining semantics, modality-specific spatial attention, and shared channel attention; \eg, results on the Nordland dataset show an absolute improvement of $32\%$
and on RobotCar AM$\to$PM an improvement of $22\%$.
Similarly, SAANE also shows improvements between $37\%$ and $38\%$ over both implementations of APANet.
Adding the semantic modality to APANet-MM is insufficient to significantly boost the performance of this pooling method in this scenario.

\section{Conclusion}

We present an attention-based, semantically-aware deep embedding model for image-based visual localization: SAANE.  Our
model targets the sensitivity of appearance-based visual localization to both perceptual aliasing from repetitive
structures and extreme differences in viewing conditions.
SAANE uses a novel multimodal attention module that fuses both mid-level appearance and high-level semantic features by
predicting attention maps for both modalities with a shared channel attention.  The attended maps are then fused to
generate semantically informed image embedding.  We evaluate the performance benefits of our model on three challenging
public visual localization datasets, while showing significant gains compared to the baseline ($12\%$) and prior
state-of-the-art methods ($19\%$).  We also show
that SAANE learns to produce stable attention maps that focus on consistent image regions across changing views,
which is crucial for autonomous navigation.

\section{Acknowledgement}
We would like to thank Ajay Divakaran for proof-reading the manuscript and providing helpful comments. 

\bibliography{biblio}

\clearpage

\section{Implementation Details}
\label{appsec:impl}
\begin{figure}[ht]
\centering
\resizebox{0.6\textwidth}{0.4\textwidth}{\input{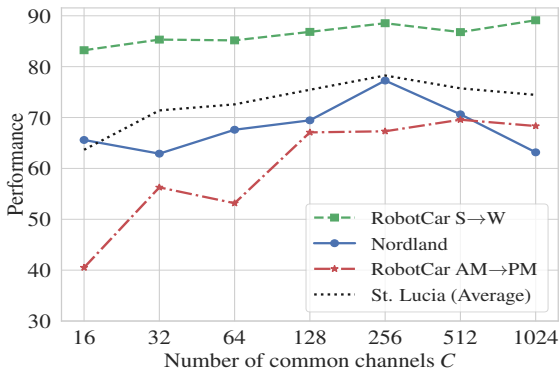}}
\caption{Figure shows a plot of performance of SAANE versus the number of channels $C$ in the 
common multimodal fusion space.}
\label{fig:channels}
\end{figure}

The backbone of SAANE is two parallel DCNNs.  We use a ResNet50~\cite{he2016deep} pre-trained for the Imagenet classification task for mid-level feature extraction and a Pyramid Scene Parsing Network
(PSPNet)~\cite{zhao17:pspnet} pre-trained on the ADE20K~\cite{zhou2017scene} semantic segmentation task for extracting
semantic features. We also experimented with a version of PSPNet pre-trained on Cityscapes~\cite{Cordts2016Cityscapes};
however, we found the ADE20K version to be more robust to viewpoint changes and the 150 classes of ADE20K to be more useful in the presence of diverse scene types.  We use the output of the third residual block from ResNet50 as mid-level appearance features
($\mF^{A}$).  For high-level semantic features ($\mF^{S}$), we use the output before the final convolutional layer of
PSPNet. The resulting number of channels in appearance and semantic features are $C_R = 1024$ and $C_S = 512$,
respectively. 
We set the number of channels of the common embedding space in the modality fusion module ($C$), both before and after MM-Att, to $256$.
We use spatial pyramid pooling with pooling sizes of $[4, 3, 2, 1]$ and concatenate the feature maps from all layers to produce the final embedding.
The dimensionality of the final embeddings after spatial pooling is $7680$.  
For our experiments, we the two pre-trained DCNNs and 
fine-tune the two modality fusion modules and the MM-Att module.
We use the Adam optimizer~\cite{kingma2014adam} with a learning rate of $5 \times
10^{-5}$ and weight decay of $5 \times 10^{-4}$ for training.  We use online triplet sampling with batches comprised of 16 different
classes with 4 examples per class.  Within a batch, we utilize distance-weighted triplet
sampling~\cite{manmatha17:sampling-matters} to increase the stability of the training process. 
We use a margin of $m=0.5$, selected based on giving the best performance on a small validation set.
Due to our assumption that our test data come from a dissimilar distribution as our training data, we did not by default experiment with any form of whitening as used in~\cite{arandjelovic18:netvlad,zhu18:apa}. 

Finally, to explore the effect of the model capacity on performance on the test datasets, we experiment with varying the dimensionality $C$ of the multimodal fusion network.
As shown in \autoref{fig:channels}, the performance across all of the datasets plateaus between $128$ and $256$ channels and shows evidence of overfitting, particularly in the case of Nordland, above $256$ channels.
The best dimensionality of the multimodal fusion module also appears to be a function of the dataset difficulty.
Our model's performance on RobotCar Sum.$\to$Win., in the presence of minor seasonal variations, is relatively stable, even down to $16$ channels, while the tasks with more extreme variation, such as RobotCar AM$\to$PM, sharply decline below $128$.

\section{Test Datasets}
\label{appsec:ds}
\paragraph{Nordland}~\cite{neubert2015superpixel} is derived from continuous video footage of a train journey recorded for a Norwegian television program, recorded from the front car of the train across four different seasons. 
We extract one frame per second from the first hour of each traversal, excluding images where the train is either stopped or in tunnels. 
This results in $1403$ frames per traversal. 
We perform our experiments by constructing a database with the summer traversal and querying it with the winter traversal (Sum.$\to$Win.). 
The images feature no viewpoint variation, due to travel on fixed rails; however, the seasonal appearance changes are quite severe.

\paragraph{St. Lucia}~\cite{glover10:fabmapratslam} comprises ten different traversals recorded by a forward-facing webcam affixed to the roof of a car, following a single route through the suburb of St. Lucia, Queensland, Australia.
 This dataset was captured at five different times of day on different days across two weeks.
 We use the first traversal (`100909\_0845') as the database and query with the remaining nine, reporting the average as well as the worst case result over the nine trials.
We sample images at one frame per second, which results in each traversal containing on an average $1350$ frames. 
The dataset features slight viewpoint variations due to differences in the route taken by the vehicle.  
There are mild to moderate appearance changes due to differences in time of day and the presence of dynamic objects in the scene.

\paragraph{Oxford RobotCar}~\cite{maddern20171} comprises several different traversals of the city of Oxford by a vehicle.
It was collected across varying weather conditions, seasons, and times of day over a period of a year. 
We select two pairs of traversals, referred to as Overcast Autumn/Night Autumn and Overcast Summer/Snow Winter.\footnote{The first three were introduced in~\cite{garg18:lost}, and the traversals were originally referred to in~\cite{maddern20171} as 2014-12-09-13-21-02, 2014-12-10-18-10-50, 2015-05-19-14-06-38, and 2015-02-03-08-45-10, respectively.}
 We perform an experiment by building a database with either Overcast Summer or Overcast Autumn and querying it with Snow Winter (Sum.$\to$Win.) or Night Autumn (AM$\to$PM), respectively.
  We make use of the center image from the front-facing camera and extract one frame per second.  On average, each traversal covers nearly $9$ km and $2000$ frames. 
There are mild viewpoint variations present, again due to slight differences in the starting point and road position of the traversals.  
The appearance change in the day-night pair is quite drastic, largely from the difference in illumination quality in the transition from sunlight to street lights; while it is more moderate in the summer-winter pair, with minor variation from seasonal vegetation and ground cover.

\end{document}